\documentclass[letterpaper, 10 pt, conference]{ieeeconf}  

\IEEEoverridecommandlockouts                              

\usepackage{graphics} 
\usepackage{epsfig} 
\usepackage{mathptmx} 
\usepackage{times} 
\usepackage{amsmath} 
\usepackage{amssymb}  
\usepackage{multirow}
\usepackage{cite}

\usepackage{flushend}
\usepackage{booktabs}
\usepackage{color}
\usepackage{soul}
\sethlcolor{red}

\title{\LARGE \bf
Multiple Update Particle Filter: Position Estimation by Combining GNSS Pseudorange and Carrier Phase Observations
}

\author{Taro Suzuki$^{1}$
\thanks{$^{1}$Taro Suzuki is with Future Robotics Technology Center, Chiba Institute of Technology, Japan
        {\tt\small taro@furo.org}}%
}
\begin{document}

\maketitle
\thispagestyle{empty}
\pagestyle{empty}

\begin{abstract}
This paper presents an efficient method for updating particles in a particle filter (PF) to address the position estimation problem when dealing with sharp-peaked likelihood functions derived from multiple observations. Sharp-peaked likelihood functions commonly arise from millimeter-accurate distance observations of carrier phases in the global navigation satellite system (GNSS). However, when such likelihood functions are used for particle weight updates, the absence of particles within the peaks leads to all particle weights becoming zero.  To overcome this problem, in this study, a straightforward and effective approach is introduced for updating particles when dealing with sharp-peaked likelihood functions obtained from multiple observations. The proposed method, termed as the multiple update PF, leverages prior knowledge regarding the spread of distribution for each likelihood function and conducts weight updates and resampling iteratively in the particle update process, prioritizing the likelihood function spreads. Experimental results demonstrate the efficacy of our proposed method, particularly when applied to position estimation utilizing GNSS pseudorange and carrier phase observations. The multiple update PF exhibits faster convergence with fewer particles when compared to the conventional PF. Moreover, vehicle position estimation experiments conducted in urban environments reveal that the proposed method outperforms conventional GNSS positioning techniques, yielding more accurate position estimates.
\end{abstract}

\section{Introduction}
State estimation methods, utilizing particle filters (PFs), have gained significant traction across various domains owing to their adaptability to nonlinear systems and their ability to handle multimodal distributions \cite{pf_survey0}. Particularly in robotics, PFs find widespread use in estimating the position of robots, often referred to as Monte Carlo localization (MCL) \cite{MCL_general}. Despite their effectiveness, PFs encounter challenges, notably when the distribution of particles diverges from the likelihood function obtained from observations, resulting in zero weights for all particles. This issue is especially pronounced when the likelihood function exhibits a sharp peak within a narrow distribution concerning the state variables, when the initial particle distribution is widely spread, or when the number of particles is relatively small compared to the state's dimensionality. To mitigate these challenges, the annealed PF approach has been introduced \cite{annealed_pf1}. This method aims to smoothen the likelihood function through observations and iteratively adjust the particle distribution over multiple steps. However, its applicability is limited as it assumes a single observation for computing the likelihood function, rendering it ineffective for scenarios where multiple observations yield likelihood functions with varying distribution shapes.

In the context of particle filters, when multiple independent observations are acquired, the likelihood of particles is commonly computed as the product of these individual likelihood functions. However, as depicted in Fig. 1, a significant issue arises if one of these observations yields a likelihood function characterized by a sharp peak with a narrow distribution concerning the state variables, as previously described. In such cases, insufficient particle density leads to the weights not being updated. Moreover, when the likelihood function exhibits multimodal sharp peaks, the particles may converge to local minima, hindering accurate estimation. Although the cascade PF approach has been proposed to address these challenges by conducting multiple weight updates and resampling from the likelihood functions obtained from various observations \cite{cascade_pf}, it fails to consider the breadth and shape of the distribution of likelihood functions associated with each observation.

\begin{figure}[t!]
   \centering
   \includegraphics[width=80mm]{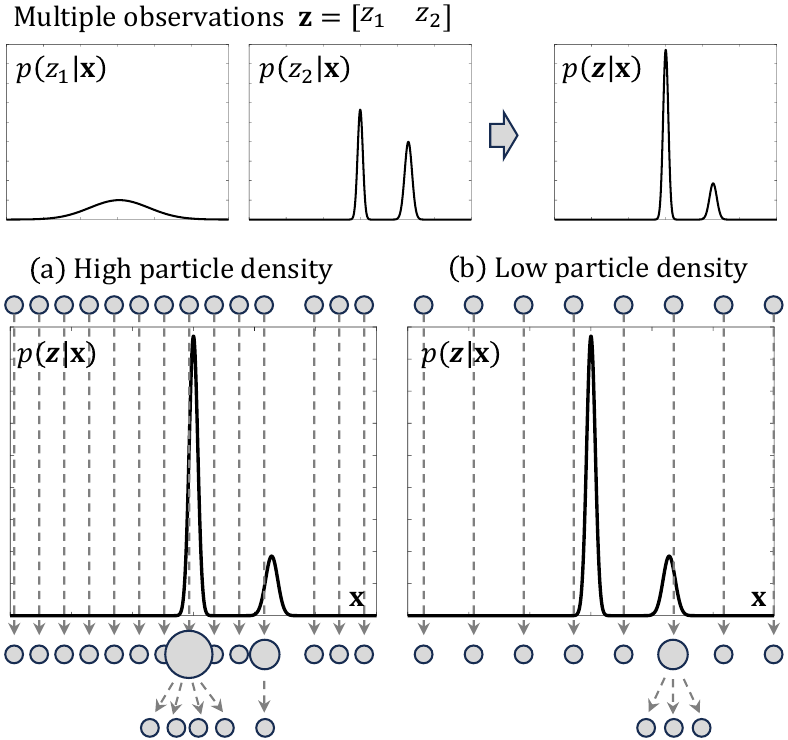} 
   \caption{ Illustration of the normal particle filter (PF) in the presence of multiple observations with sharp peaks. When the initial particle distribution is widely spread and the number of particles is limited, the weights of particles at their true positions are not evaluated.}
   \label{fig1}
\end{figure}

In outdoor mobile robot localization, position estimation via the global navigation satellite system (GNSS) is widely employed. GNSS carrier phase observations offer precise distance information down to the millimeter level, indicating the distance between the satellite and receiving antenna. However, these measurements are subject to integer wavelength ambiguities, resulting in multiple local solutions represented by a grid of wavelength sizes. Conversely, GNSS pseudorange observations provide distance information without ambiguity, albeit with an accuracy at the meter level. When incorporating these GNSS observations into a PF, challenges arise. The likelihood function derived from carrier phase observations exhibits multiple sharp peaks at the centimeter level. Depending on the prior distribution of particles, this can lead to scenarios where the weights of all particles become zero or where particles converge to a local minimum.

Therefore, assuming that multiple observations exist and that the spread of the distribution of each likelihood function is available as prior knowledge, we propose a multiple update PF (MU-PF). This method performs multiple weight updates and resamplings in the order of the spread of the distribution of the likelihood functions in the particle update process.

The contributions of this study are as follows.
\begin{itemize}
   \item The proposed MU-PF considers the distribution spread of multiple observations, enabling accurate state estimation even in scenarios where the previously suggested PF fails due to uniform particle weights in the observation updates.
   \item The development of a novel likelihood function utilizing GNSS pseudorange and carrier phase observations, and its integration into the MU-PF framework to enhance convergence performance and improve position estimation accuracy in comparison to conventional methods.
\end{itemize}

\section{Related Studies}
Numerous studies have focused on enhancing PF, introducing a range of strategies to enhance accuracy, convergence, and computational efficiency \cite{pf_general1,pf_general2}. In particular, many methods have been studied to correct the bias of the particle distribution for a problem termed as sample degeneracy and impoverishment \cite{pf_survey1}.Among the challenges encountered in PF, a notable issue arises when the likelihood distribution of particles derived from observations exhibits an exceedingly sharp peak, leading to particle degeneracy when the particle count is insufficient relative to the state dimensions. To address this, the Annealed PF approach \cite{annealed_pf1} (or annealed importance sampling \cite{annealed_is}) has been proposed as an effective remedy. Annealed PF leverages annealing techniques to smooth the likelihood function, thereby elucidating the global maximum and optimizing particle distribution via multi-stage sampling. Additionally, a method integrating the kernel mean shift algorithm with Annealed PF has been proposed to further enhance particle distribution quality \cite{annealed_pf0}. However, Annealed PF's applicability is limited in scenarios involving independent observations from multiple observers.

Conversely, in PFs, where multiple observations are involved, particle weights are typically computed as a single observer by the multiplication of likelihood functions. An alternative approach, termed as Partitioning PF \cite{partitioned_pf}, has been proposed to address multiple observations independently. This method partitions the state space into multiple segments and treats each segment individually by sampling with weights determined by different likelihood functions. Although effective in reducing the required number of particles for managing high-dimensional state spaces, such approaches are less suitable for 3D position estimation. Another method, Cascade PF \cite{cascade_pf}, has been introduced to perform multiple resamplings from distinct independent observations. Although similar in concept to the proposed method, Cascade PF does not specify the sequence in which the multiple observations are applied. 

On the other hand, methods such as MCL that integrate GNSS positioning solutions with lidar \cite{MCL_GPS1, MCL_GPS2} and combine odometry with inertial sensors \cite{MCL_GPS3} have been proposed. They showcase the utilization of GNSS observations within the PF framework based on loosely coupled integration. 
In previous studies, such as \cite{taro_gnss_pf1} and \cite{taro_gnss_pf2}, mobile robot positioning has been estimated utilizing GNSS pseudorange observations based on tightly coupled integration. Nevertheless, these studies do not incorporate carrier phase observations and are limited to meter-level accuracy in position estimation. When employing carrier phase observations, the estimation of integer ambiguity becomes necessary. Methods, such as the ambiguity function method (AFM) \cite{AFM} and its improved variations, modified AFM (MFAM) \cite{MFAM1,MFAM2,MFAM3}, have been proposed for this purpose. In our study, we specifically focus on the ambiguity function value (AFV), which serves as the objective function in the optimization process for AFM. Leveraging AFV, we compute particle weights to achieve centimeter-level accuracy in position estimation without the need for carrier phase ambiguity estimation.

\section{Proposed Method}
\subsection{Multiple Update Particle Filter}
The objective of PF is to estimate $p\left(\mathbf{x}_t \mid \mathbf{Z}_t\right)$, which represents the distribution of target states $\mathbf{x}_t$ at time $t$ given all observations $\mathbf{Z}_t = (\mathbf{z}_1, \ldots ,\mathbf{z}_t)$. PF can be described in the following two steps.

(1) Prediction
\begin{equation}
   p\left(\mathbf{x}_t \mid \mathbf{Z}_{t-1}\right)=\int p\left(\mathbf{x}_t \mid \mathbf{x}_{t-1}\right) p\left(\mathbf{x}_{t-1} \mid \mathbf{Z}_{t-1}\right) d \mathbf{x}_{t-1}
\end{equation}

(2) Update or correction
\begin{equation}
   p\left(\mathbf{x}_t \mid \mathbf{Z}_t\right) \propto p\left(\mathbf{z}_t \mid \mathbf{x}_t\right) p\left(\mathbf{x}_t \mid \mathbf{Z}_{t-1}\right)
\end{equation}

The computation of the integral in the above equation is approximated by importance sampling \cite{pf_general2}. When multiple observations $\mathbf{z}=(z_1,\ldots,z_M)$ are obtained from multiple observers, the likelihood function is expressed as the product of the likelihoods from each observation, assuming that each observation is independent.

\begin{equation}
   p(\mathbf{z} \mid \mathbf{x})=p\left(z_1, \ldots, z_M \mid \mathbf{x}\right)=\prod_{m=1}^M p\left(z_m \mid \mathbf{x}\right)
\end{equation}

\noindent where $M$ denotes the number of observers. In normal PF, the particle weights are updated and resampled from the likelihood function shown in (3), which combines multiple observations. However, in (3), if one of the observations exhibits a sharp peak with a narrow distribution, then the observation dominates. Hence, correct state estimation is not possible when the number of particles is small, as shown in Fig. 1.

Fig. 2 shows the operation of the proposed MU-PF. Specifically, MU-PF repeats particle likelihood and resampling in multiple steps, applying the individual likelihood functions obtained from multiple observations in the appropriate order. For notational simplicity, we denote the prior distribution before the observation update at time $t$ as follows:

\begin{equation}
   \pi_0\left(\mathbf{x}_t\right)=p\left(\mathbf{x}_t \mid \mathbf{Z}_{t-1}\right)
\end{equation}

Consider the intermediate distribution of targets at update in multiple steps. At $m$-th step, $\pi_{m-1}\left(\mathbf{x}_t\right)$ is used as the proposed distribution of importance sampling to simulate $\pi_m\left(\mathbf{x}_t\right)$ with weighted particles. 

\begin{equation}
   \pi_m\left(\mathbf{x}_t\right)=p\left(z_{m, t} \mid \mathbf{x}_t\right) \pi_{m-1}\left(\mathbf{x}_t\right)
\end{equation}

The target distribution after applying multiple updates from all the observations is as follows:

\begin{equation}
   \pi_M\left(\mathbf{x}_t\right)=p\left(\mathbf{x}_t \mid \mathbf{Z}_{t-1}\right) \prod_{m=1}^M p\left(z_m,t \mid \mathbf{x}_t\right)
\end{equation}

Although a similar multiple update discussion is reported in \cite{cascade_pf}, in this study, we determine the order of the observations for a total of $M$ such that the following equation is satisfied.

\begin{equation}
   \sigma_{1}>\sigma_{2}\ldots>\sigma_{m}\ldots>\sigma_{M}
\end{equation}

\noindent where $\sigma_m$ denotes the standard deviation of the distribution of $m$-th observation when approximated by a normal distribution. The proposed method assumes that the spread of the likelihood distribution of multiple observations is known a priori. For example, in GNSS observations, the distribution spread is determined by the accuracy of the distance measurement for a given satellite configuration. Hence, the pseudorange observation exhibits the largest spread. This is followed by the carrier phase observation. By applying the pseudorange observations first and then weighting and resampling, the particles can be gradually shifted to their true positions, allowing the particle weights to be updated even if there are sharp peaks in the carrier phase observations.

Cascade PF does not consider the order in which updates from observations are applied. However, if the likelihood from observations with sharp peaks are applied first, then the particle weights will not be updated and the PF may fail. Annealing PF effectively mitigates the impact of likelihood functions with sharp peaks. Nevertheless, this adjustment comes at the cost of reduced convergence speed, as likelihood distributions with significant spread contribute minimally to updating particle weights.

\begin{figure}[t!]
   \centering
   \includegraphics[width=60mm]{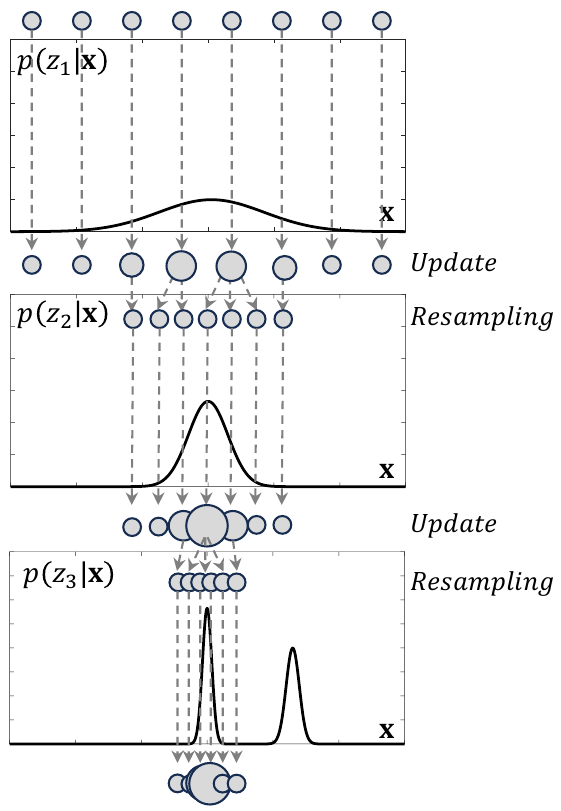} 
   \caption{Illustration of the proposed multiple update PF. By repeating weight updates and resampling in the order of the spread of the distribution of the likelihood function of multiple observations, the particles are gradually moved to the correct state.}
   \label{fig2}
\end{figure}

\begin{figure*}[t!]
   \centering
   \includegraphics[width=135mm]{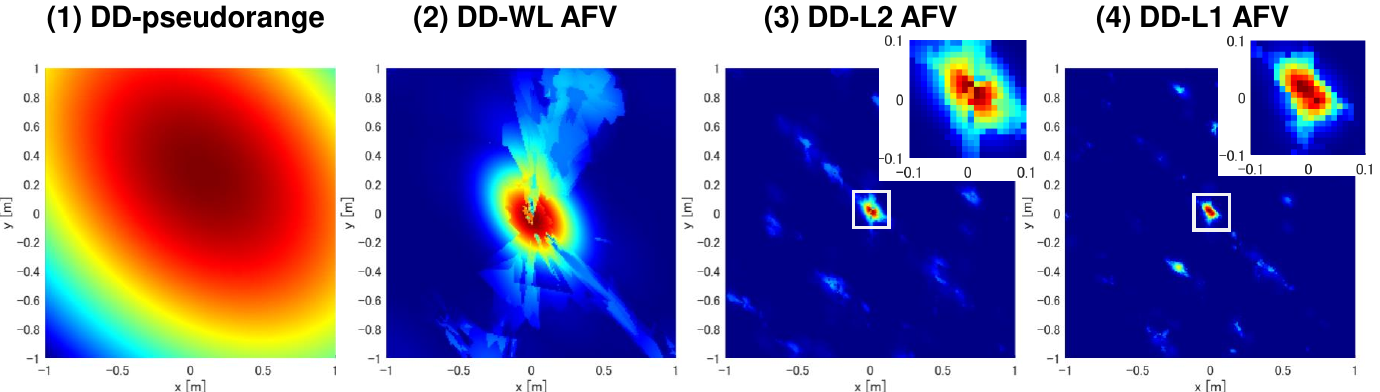} 
   \caption{Examples of likelihood computed from GNSS observations. (1) DD-pseudorange, (2) DD-WL AFV, (3) DD-L2 AFV, and (4) DD-L1 AFV. The distributions exhibit sharp peaks in this order. The likelihood from the L1 and L2 AFVs has many local maxima in the ±1 m range.}
   \label{fig3}
\end{figure*}

\subsection{Likelihood Estimation by GNSS Pseudorange and Carrier Phase}
\subsubsection{Pseudorange}
GNSS pseudorange observations include satellite orbit and clock errors, ionospheric and tropospheric delays, and receiver clock errors. However, these errors can be eliminated by double differencing (DD) between base station GNSS observations and between satellites. Let $K$ be the total number of DD-pseudorange observations. Specifically, DD-pseudorange observations can be expressed as $\mathbf{P} = (\rho^1, ... ,\rho^K)$. The residual of the $k$-th DD-pseudorange $d(\rho^k,\mathbf{x})$ at particle $\mathbf{x}$ is expressed by the following equation.

\begin{equation}
   d(\rho^k,\mathbf{x}) = \rho^k-r^k(\mathbf{x})
\end{equation}

\noindent where $\rho^k$ denotes $k$-th DD-pseudorange and $r^k(\mathbf{x})$ denotes DD-geometric distance (Euclidean distance) between the satellite and particle calculated from the particle's 3D position. If the position of the particle is the true position and there is no error in the DD-pseudorange observation, then (8) becomes zero. The likelihood of a particle by the DD-pseudorange $p\left(\rho^k \mid \mathbf{x}\right)$ can be calculated as follows:

\begin{equation}
   p\left(\rho^k \mid \mathbf{x}\right)=\frac{1}{\sqrt{2 \pi} \sigma_\rho} \exp \left(-\frac{d(\rho^k,\mathbf{x})^2}{2 \sigma_\rho^2}\right)
\end{equation}

\noindent $\sigma_\rho$ is the observed accuracy of the DD-pseudorange, which is typically approximately 0.5 m. Combining the DD-pseudorange of all satellites, the likelihood of a particle by the DD-pseudorange is calculated by the following equation.

\begin{equation}
   p\left(\mathbf{P} \mid \mathbf{x}\right)=\prod_{k=1}^K p\left(\rho^k \mid \mathbf{x}\right)
\end{equation}

\subsubsection{Carrier Phase}
With respect to position estimation using the GNSS carrier phase, RTK-GNSS , which employs the DD-carrier phase between the base station and satellite, is commonly used \cite{gnss_enge}. In RTK-GNSS, realizing centimeter-accurate solutions typically involves estimating integer ambiguity in carrier phase observations through integer least squares \cite{lambda}. However, in this study, a method to calculate the likelihood of particles is proposed by directly using carrier phase observations. We utilize the AFV to compute particle weights. Specifically, AFM is a method that searches for integer ambiguities of the carrier phase in 3D space. Furthermore, AFM uses AFV as the objective function and searches for the integer ambiguity that minimizes it. Instead of using AFV to estimate the carrier phase ambiguity, we propose a method to determine the particle weights by using the AFV calculated from the carrier phase observation.

Let $\boldsymbol{\Phi} = (\Phi^1, ... ,\Phi^K)$ be the DD-carrier phase observation. The AFV $\psi$ computed from $k$-th DD-carrier phase $\Phi^k$ can be expressed by the following equation \cite{MFAM1}.

\begin{equation}
   \psi\left(\Phi^k, \mathbf{x}\right)=\operatorname{round}\left(\Phi^k-\frac{1}{\lambda} r\left(\mathbf{x}\right)\right)-\left(\Phi^k-\frac{1}{\lambda} r\left(\mathbf{x}\right)\right)
\end{equation}

where $\lambda$ denotes the wavelength of carrier wave. In this equation, the AFV exhibits an exceedingly large number of minima over an interval of wavelength $\lambda$. Therefore, combining carrier phase observations from multiple satellites can reduce local peaks in the likelihood. The likelihood of particles by AFV can be calculated using the following equation.

\begin{equation}
   p\left(\Phi^k \mid \mathbf{x}\right)=\frac{1}{\sqrt{2 \pi} \sigma_\Phi} \exp \left(-\frac{\psi\left(\Phi^k, \mathbf{x}\right)^2}{2 \sigma_\Phi^2}\right)
\end{equation}

The combined likelihood of AFVs from all satellites can be expressed as follows:

\begin{equation}
   p\left(\boldsymbol{\Phi} \mid \mathbf{x}\right)=\prod_{k=1}^K p\left(\Phi^k \mid \mathbf{x}\right)
\end{equation}

GNSS transmits signals at multiple frequencies, and carrier phase observations can be obtained at several different wavelengths. In this study, we use carrier phase observations at two frequencies, L1 (1.57542 GHz: 19 cm wavelength) and L2 (1.2276 GHz: 25 cm wavelength). By linearly combining the carrier phase observations, it is possible to generate phase observations with pseudo-different wavelengths. In this case, the L1-L2 carrier phase observation is used, which is termed as the wide-lane (WL) linear combination \cite{gnss_handbook}.

\begin{equation}
   \boldsymbol{\Phi_{WL}}=\boldsymbol{\Phi_{L1}}-\boldsymbol{\Phi_{L2}}
\end{equation}

Here, the wavelength of the WL carrier phase is approximately 80 cm. Furthermore, given that the ranging performance of the carrier phase also depends on the wavelength, the following equation is obtained.

\begin{equation}
   \sigma_{\rho}>\sigma_{\Phi_{WL}}>\sigma_{\Phi_{L2}}>\sigma_{\Phi_{L1}}
\end{equation}

Therefore, as shown in (15), the observations are applied to the MU-PF in the order of 1. DD-pseudorange, 2. DD-WL carrier phase, 3. DD-L2 carrier phase, and 4. DD-L1 carrier phase.

\subsubsection{Examples of Likelihood Estimation}
Fig. 3 shows an example of calculating weights at points on the grid centered on the true position using (10) and (13) from actual GNSS observations. Here, the grid size is 1 cm, and the grid range is ±1 m from the true position. In Fig. 3, the DD-pseudorange exhibits the widest range with a unimodal distribution. Conversely, the WL carrier phase exhibits a wavelength of 80 cm. This yields a single peak that can be identified in the ±1 m range. However, the weights calculated by AFV from the L1 and L2 observations exhibit multiple sharp peaks. By applying particle weighting and resampling in the order shown in Fig. 3, the particle weights do not become zero even if the initial particle distribution is widespread. Hence, the density of particles at the correct location increases, allowing position estimation.

\begin{figure*}[t!]
   \centering
   \includegraphics[width=165mm]{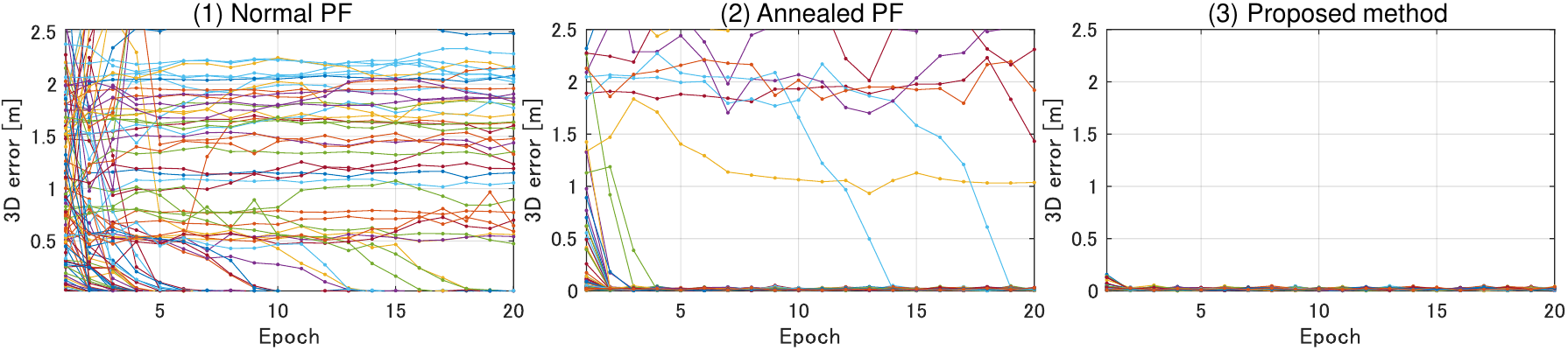} 
   \caption{Comparison of 3D position estimation errors over 100 trials using different GNSS observations. The proposed method converges to within 10 cm of the 3D position estimation error after almost one observation when compared to other methods.}
   \label{fig4}
\end{figure*}

\section{Evaluation by Static Test}
We evaluate the performance of the proposed position estimation method by employing MU-PF using GNSS observations acquired in a static environment. Our evaluation involves comparing the proposed method with the normal PF, an annealed PF \cite{annealed_pf1}. All methods utilize the same parameters except for the manner in which the likelihood function is applied and the update process. Multinomial resampling is employed for all comparison methods. 
In the case of Annealed PF, the number of annealing stages is identical to that of the proposed method ($M=4$). The annealing rate is determined empirically. The initial particle distribution is generated around the true position, with a normal distribution of ±2 m in the $x$, $y$, and $z$ axes, respectively. 


\subsection{Evaluation of convergence}
One hundred sets of 20 epochs (20 s) of GNSS observations, starting at different times within 1-h GNSS observations, are prepared. Furthermore, position estimation is performed using PF to evaluate the convergence of each method. The number of particles was set to $N=2000$. 

Fig. 4 illustrates the 3D position estimation error at each epoch for each method, while Table 1 presents the 3D error at the first and last epochs (after 20 s) for each method, along with the percentage of positions estimated within 10 cm. It is evident that the normal PF fails to converge to the correct position in many trials, even after 20 s of observation updates. This failure can be attributed to the presence of a sharp peak in the likelihood function, resulting in the weight of numerous particles becoming zero due to the product of likelihood functions derived from pseudorange and carrier phase data. Although the annealed PF demonstrates particle convergence toward the true position as time progresses, some trials converge to local minima. Conversely, the proposed method achieves convergence to the correct position in nearly one observation step. This rapid convergence is facilitated by applying likelihood functions in the correct order and performing resampling, allowing particles to gradually shift and ultimately converge to the peak position of the likelihood function derived from L1 carrier phase observations.

After 20 observation updates, the normal PF achieves 3D position estimates with a 10-cm accuracy in 50 \% of the trials out of 200, while the annealed PF achieves this in 93 \% of trials. Remarkably, the proposed method achieves a 100 \% success rate in accurate 3D position estimation. These experiments underscore the superiority of the proposed method in accurately estimating positions, especially in scenarios where the normal PF fails. Moreover, the proposed method demonstrates faster convergence compared to the annealed PF and realizes centimeter-level accuracy with just a single GNSS observation.

\begin{table}[]
   \caption{Performance evaluation of the proposed method.}
   \label{tab:tab1}
   \label{tab:my-table}
   \resizebox{\columnwidth}{!}{%
   \begin{tabular}{@{}c|cc|cc@{}}
   \toprule
   \multirow{2}{*}{Method} & \multicolumn{2}{c|}{Epoch: 1}                & \multicolumn{2}{c}{Epoch: 20}                \\ \cmidrule(l){2-5} 
                           & 3D error {[}cm{]} & Fixed rate {[}\%{]} & 3D error {[}cm{]} & Fixed rate {[}\%{]} \\ \midrule
   Normal PF               & 111.44            & 8.0                 & 89.39             & 50.0                \\
   Annealed PF \cite{annealed_pf1}             & 38.55             & 67.0                & 18.63             & 93.0                \\
   Proposed method         & \textbf{6.89}     & \textbf{96.0}       & \textbf{1.64}     & \textbf{100}        \\ \bottomrule
   \end{tabular}%
   }
\end{table}


\begin{figure}[t!]
   \centering
   \includegraphics[width=80mm]{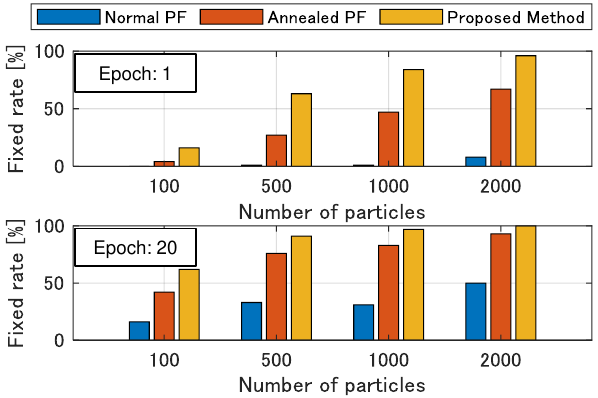} 
   \caption{Percentage of 3D position estimation within 10 cm for each method when the number of particles is varied. 
   }
   \label{fig5}
\end{figure}

\subsection{Evaluation based on number of particles}
We evaluate the performance of each method when the number of particles is reduced. Furthermore, we utilize the same dataset as employed for the convergence evaluation, varying the number of particles at 100, 500, and 1000. The initial particle distribution remains consistent, generated with a normal distribution of ±2 m.

Fig. 5 illustrates the percentage of the 100 trials accurately estimated within 10 cm after the first observation update ($t=1$) and after 20 observation updates ($t=20$). The convergence performance of the proposed method deteriorates when the number of particles is insufficiently dense for the state space. For instance, with 100 particles, 62 \% of the trials achieved position estimates within 10 cm after 20 GNSS observations. However, when compared to the annealed PF, the proposed method exhibits less degradation in position estimation performance despite the decrease in the number of particles. This resilience is attributed to the systematic application of likelihood functions from observations with varying distribution sizes in a suitable order, facilitating a gradual transition of particles. Moreover, a distinguishing feature of the proposed method is its heightened probability of achieving position estimation performance within 10 cm in a single observation, even with a reduced number of particles. The experimental findings underscore that the proposed method can estimate the state with fewer particles than other methods, thereby reducing computational costs while maintaining comparable convergence performance.

\section{Evaluation based on Kinematic Test}
In the kinematic test, the performance of the proposed method in position estimation is evaluated in contrast to conventional RTK-GNSS by estimating the position of a vehicle in a real-world setting. For the evaluation, we utilize the UrbanNav Tokyo dataset \cite{urbannav}, which is publicly available as open data. This dataset comprises GNSS observation data recorded from a vehicle traversing an urban area, along with GNSS observation data collected from a base station, and reference vehicle positions for evaluation purposes.

The dataset showcases driving routes and surrounding environment as depicted in Fig. 6. The color of the trajectory in the figure corresponds to the number of GNSS satellites received. Certain areas feature structures and buildings that obstruct GNSS signals, leading to multipath errors. As a comparison method, RTKLIB \cite{rtklib}, a well-established open-source GNSS positioning library, is employed. RTKLIB operates in kinematic mode with a satellite elevation angle mask set at 15° and a signal strength mask set at 35 dB-Hz. Default settings are maintained for all other parameters. In the kinematic test, the predictive model of the PF incorporates the estimated 3D velocity derived from GNSS Doppler observations.

Fig. 7 displays the time-series 3D position error of the positioning solution obtained using the proposed method and RTKLIB. In environments where the number of available GNSS satellites is diminished due to obstruction, the positioning error of both methods tends to increase. However, the proposed method exhibits a more moderate increase in positioning error overall. Furthermore, Fig. 8 illustrates the cumulative distribution function of the 3D position error. The percentage of position estimation accuracy within 0.5 m for the conventional RTK-GNSS method was 63.9 \%, while it increased to 80.3 \% for the proposed method. These results underscore that the proposed method, leveraging MU-PF for GNSS pseudorange and carrier phase observations, can realize more accurate position estimation when compared to conventional RTK-GNSS method.

\begin{figure}[t!]
   \centering
   \includegraphics[width=80mm]{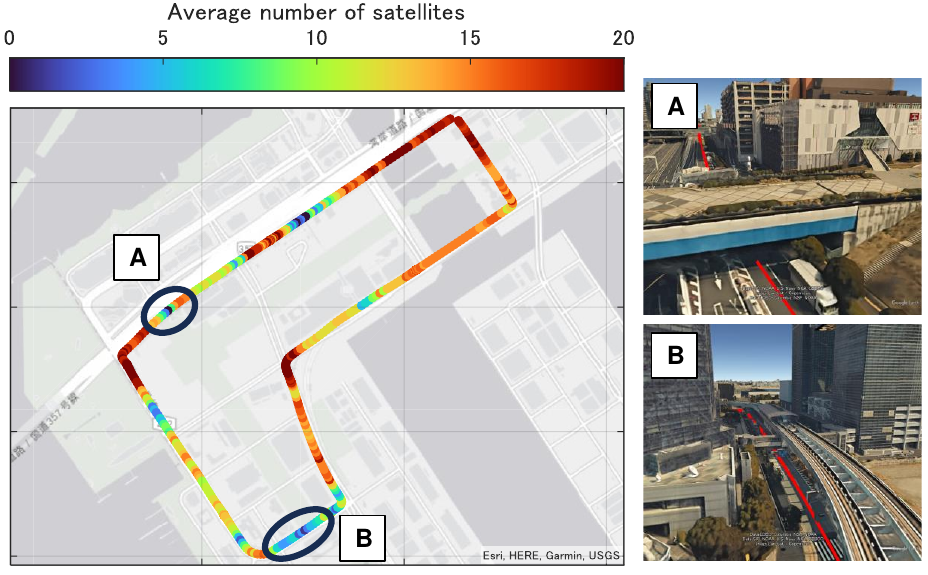} 
   \caption{Trajectory and environment of the kinematic test.}
   \label{fig6}
\end{figure}
\begin{figure}[t!]
   \centering
   \includegraphics[width=85mm]{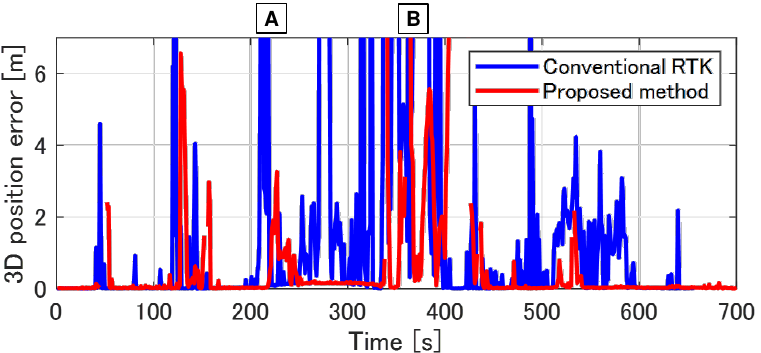} 
   \caption{Comparison of 3D position estimation error between the proposed method and conventional RTK-GNSS.}
   \label{fig7}
\end{figure}
\begin{figure}[t!]
   \centering
   \includegraphics[width=85mm]{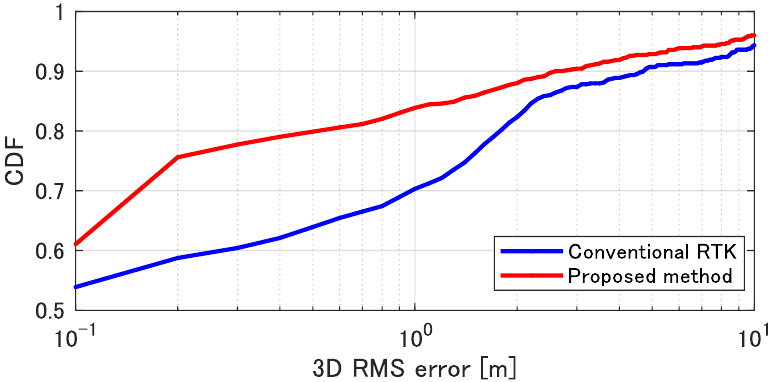} 
   \caption{Cumulative distribution function of position estimation error by the proposed method in the kinematic test and comparison with an RTK-GNSS.}
   \label{fig8}
\end{figure}
\section{Conclusion}
In this study, an efficient particle update method is proposed for state estimation problems in PF when multiple observations are available, and one of these observations yields a sharply peaked likelihood function. Operating under the assumption that the broad distribution of particle likelihood functions derived from observations is known, we proposed an MU-PF. This approach iteratively updates and resamples weights in accordance with the broad distribution of likelihood functions. In static experiments utilizing GNSS pseudorange and carrier phase observations, 
the proposed method correctly applies pseudorange and carrier phase observations in the appropriate order, enabling the PF to accurately estimate the correct position. Furthermore, comparative experiments demonstrate that the proposed method exhibits superior convergence performance with fewer particles than the annealed PF. Additionally, in a position estimation experiment involving a vehicle navigating an urban environment, the proposed method demonstrates accurate position estimation when compared to conventional RTK-GNSS approaches.

The likelihood functions for all observations were Gaussian in this study. Hence, as future research, we will consider integration with sensors that consider different distributions such as Lidar and cameras. We also plan to examine the application of the proposed method to the case involving multiple observations with multimodal distributions.

\addtolength{\textheight}{0cm}   


\bibliographystyle{IEEEtran}
\bibliography{IEEEabrv,ICRA2024}

\end{document}